\documentclass[11pt,a4paper]{article}
\usepackage{times,latexsym}
\usepackage{url}
\usepackage[T1]{fontenc}

\usepackage[acceptedWithA]{tacl2018v2}

\usepackage{amsmath}
\usepackage{amsthm}
\usepackage{booktabs}
\usepackage{multicol}
\usepackage{adjustbox}
\usepackage{subcaption}
\usepackage{multirow}
\usepackage{algpseudocode}
\usepackage{algorithm}
\usepackage{amssymb}
\usepackage{makecell}
\usepackage[multiple]{footmisc}
\usepackage{resizegather}
\usepackage{microtype}
\usepackage[font=small]{caption}  

\usepackage{float}

\usepackage{xspace,mfirstuc,tabulary}

\newif\iftaclinstructions
\taclinstructionsfalse 
\iftaclinstructions
\renewcommand{\confidential}{}
\renewcommand{\anonsubtext}{(No author info supplied here, for consistency with
TACL-submission anonymization requirements)}
\newcommand{\instr}
\fi

\iftaclpubformat 

\else

\fi

\title{ \vspace*{-0.5in}
{{\small \hfill TACL'21}\\
\vspace*{.25in}} A Statistical Analysis of Summarization Evaluation Metrics Using Resampling Methods}

\author{
    Daniel Deutsch, Rotem Dror, and Dan Roth \\
    Department of Computer and Information Science \\
    University of Pennsylvania \\
    \texttt{\{ddeutsch,rtmdrr,danroth\}@seas.upenn.edu} \\
}

\date{}

\newif\ifcomments
\commentstrue
\ifcomments
    \providecommand\dd[1]{\textcolor{blue}{[DD: #1]}}
    \providecommand\dr[1]{\textcolor{olive}{[DR: #1]}}
    \providecommand\rd[1]{\textcolor{purple}{[RD: #1]}}
    \providecommand\todo[1]{\textcolor{red}{[TODO: #1]}}
\else
    \providecommand{\dd}[1]{}
    \providecommand{\dr}[1]{}
    \providecommand{\rd}[1]{}
    \providecommand{\todo}[1]{}
\fi

\newcommand{\rsum}{r_\textsc{Sum}}
\newcommand{\rsys}{r_\textsc{Sys}}

\newcommand{\rhosum}{\rho_\textsc{Sum}}
\newcommand{\rhosys}{\rho_\textsc{Sys}}

\begin{document}

\maketitle

\begin{abstract}
The quality of a summarization evaluation metric is quantified by calculating the correlation between its scores and human annotations across a large number of summaries. Currently, it is unclear how precise these correlation estimates are, nor whether differences between two metrics' correlations reflect a true difference or if it is due to mere chance. In this work, we address these two problems by proposing methods for calculating confidence intervals and running hypothesis tests for correlations using two resampling methods, bootstrapping and permutation. After evaluating which of the proposed methods is most appropriate for summarization through two simulation experiments, we analyze the results of applying these methods to several different automatic evaluation metrics across three sets of human annotations. We find that the confidence intervals are rather wide, demonstrating high uncertainty in the reliability of automatic metrics. Further, although many metrics fail to show statistical improvements over ROUGE, two recent works, QA\-Eval and BERTScore, do in some evaluation settings.\footnote{
Our code is available at \url{https://github.com/CogComp/stat-analysis-experiments}.
}
\end{abstract}
\section{Introduction}
Accurately estimating the quality of a summary is critical for understanding whether one summarization model produces better summaries than another.
Because manually annotating summary quality is costly and time consuming, researchers have developed automatic metrics that approximate human judgments \citep[][among others]{Lin04,TratzHo08,GKVS08,ZPLGME19,DeutschBeRo20}.

Currently, automatic metrics themselves are evaluated by calculating the correlations between their scores and human-annotated quality scores.
The value of a metric's correlation represents how similar its scores are to humans', and one metric is said to be a better approximation of human judgments than another if its correlation is higher.

However, there is no standard practice in summarization for calculating confidence intervals (CIs) for the correlation values or running hypothesis tests on the difference between two metrics' correlations.
This leaves the community in doubt about how effective automatic metrics really are at replicating human judgments
as well as whether the difference between two metrics' correlations is truly reflective of one metric being better than the other or if it is an artifact of random chance.

In this work, we propose methods for calculating CIs and running hypothesis tests for summarization metrics.
After demonstrating the usefulness of our methods through a pair of simulation experiments, we then analyze the results of applying the statistical analyses to a set of summarization metrics and three datasets.

The methods we propose are based on the resampling techniques of bootstrapping \citep{EfronTi93} and permutation \citep{Noreen89}.
Resampling techniques are advantageous because, unlike parametric methods, they do not make assumptions which are invalid in the case of summarization (\S\ref{sec:fisher}; \S\ref{sec:williams}).
Bootstrapping and permutation techniques use a subroutine that samples a new dataset from the original set of observations.
Since the correlation of an evaluation metric to human judgments is a function of \emph{matrices} of values (namely the metric's scores and human annotations for multiple systems across multiple input texts; \S\ref{sec:prelim}), this subroutine must sample new \emph{matrices} in order to generate a new instance, in contrast to standard applications of bootstrapping and permutation that sample vectors of numbers. 
To that end, we propose three different bootstrapping (\S\ref{sec:ci_bootstrapping}) and permutation (\S\ref{sec:hypo_permutation}) techniques for resampling matrices, each of which makes different assumptions about whether the systems or inputs are constant or variable in the calculation.

In order to evaluate which resampling methods are most appropriate for summarization, we perform two simulations.
The first demonstrates that the bootstrapping resampling technique which assumes both the systems and inputs are variable produces CIs that generalize best to held-out data (\S\ref{sec:ci_simulations}).
The second shows that the permutation test which makes the same assumption has more statistical power than the equivalent bootstrapping method and Williams' test \citep{Williams59}, a parametric hypothesis test that is popular in machine translation (\S\ref{sec:power}).

Finally, we analyze the results of estimating CIs and applying hypothesis testing to a set of summarization metrics using annotations on English single- and multi-document datasets \citep{DangOw08,FKMSR21,BGALN20}.
We find that the CIs for the metrics' correlations are all rather wide, indicating that the summarization community has relatively low certainty in how similarly automatic metrics rank summaries with respect to humans (\S\ref{sec:ci_experiments}).
Additionally, the hypothesis tests reveal that QA\-Eval \citep{DeutschBeRo20} and BERTScore \citep{ZKWWA20} emerge as the best metrics in several of the experimental settings, whereas no other metric consistently achieves statistically better performance than ROUGE \citep[\S\ref{sec:hypo_experiments};][]{Lin04}.

Although we focus on summarization, the techniques we propose can be applied to evaluate automatic evaluation metrics in other text generation tasks, such as machine translation or structure-to-text.
The contributions of this work include
(1) a proposal of methods for calculating CIs and running hypothesis tests for summarization metrics,
(2) simulation experiments that provide evidence for which methods are most appropriate for summarization,
and (3) an analysis of the results of the statistical analyses applied to various summarization metrics on three datasets.
\section{Preliminaries: Evaluating Metrics}
\label{sec:prelim}
Summarization evaluation metrics are typically used to either argue that a summarization system generates better summaries than another or that an individual summary is better than another for the same input.
How similarly an automatic metric does these two tasks with respect to humans is quantified as follows.

Let $\mathcal{X}$ be an evaluation metric that is used to approximate some ground-truth metric $\mathcal{Z}$.
For example, $\mathcal{X}$ could be ROUGE and $\mathcal{Z}$ could be a human-annotated summary quality score.
The similarity of $\mathcal{X}$ and $\mathcal{Z}$ is evaluated by calculating two different correlation terms on a set of summaries.
First, the summaries from summarization systems $\mathcal{S} = \{S_1, \dots, S_N\}$ on input document(s) $\mathcal{D} = \{D_1, \dots, D_M\}$ are scored using $\mathcal{X}$ and $\mathcal{Z}$.
We refer to these scores as matrices $X, Z \in \mathbb{R}^{N \times M}$ in which $x_i^j$ and $z_i^j$ are the scores of $\mathcal{X}$ and $\mathcal{Z}$ on the summary output by system $S_i$ on input $D_j$.
Then, the correlation between $X$ and $Z$ is calculated at one of the following levels:
\begin{gather*}
    \rsys(X, Z) = \textsc{Corr}\left(\left\{\left(\frac{1}{M}\sum_j x^j_i, \frac{1}{M}\sum_j z^j_i\right)\right\}_{i=1}^N\right)
\end{gather*}
\begin{equation*}
    \rsum(X, Z) = \frac{1}{M} \sum_j \textsc{Corr}\left(\left\{\left(x^j_i, z^j_i\right)\right\}_{i=1}^N\right)
\end{equation*}
where $\textsc{Corr}(\cdot)$ typically calculates the Pearson, Spearman, or Kendall correlation coefficients.\footnote{
For clarity, we will refer to $\rsum$ and $\rsys$ as correlation levels and Pearson, Spearman, and Kendall as correlation coefficients.
}

These two correlations quantify how similarly $\mathcal{X}$ and $\mathcal{Z}$ score systems and individual summaries per-input for systems $\mathcal{S}$ and documents $\mathcal{D}$.
The system-level correlation $\rsys$ calculates the correlation between the scores for each system (equal to the average score across inputs), and the summary-level correlation $\rsum$ calculates an average of the correlations between the scores per-input.\footnote{
    Other definitions for the summary-level correlation have been proposed, including directly calculating the correlation between the scores for all summaries without grouping them by input document \citep{OwczarzakDa11}.
    However, the definition we use is consistent with recent work on evaluation metrics \citep{PeyrardBoGu17,ZPLGME19,BGALN20,DeutschBeRo20}
    Our work can be directly applied to other definitions as well.
}

The correlations $\rsys$ and $\rsum$ are also used to reason about whether $\mathcal{X}$ is a better approximate of $\mathcal{Z}$ than another metric $\mathcal{Y}$ is, typically by showing that $r(X, Z) > r(Y, Z)$ for either $r$.

\section{Correlation Confidence Intervals}
\label{sec:ci}
Although the strength of the relationship between $\mathcal{X}$ and $\mathcal{Z}$ on one dataset is quantified by the correlation levels $\rsys$ and $\rsum$, each $r$ is only a point estimate of the true correlation of the metrics, denoted $\rho$, on inputs and systems distributed similarly to those in $\mathcal{D}$ and in $\mathcal{S}$.
Although we cannot directly calculate $\rho$, it is possible to estimate it through a CI.

\subsection{The Fisher Transformation}
\label{sec:fisher}
The standard method for calculating a CI for a correlation is the Fisher transformation \citep{Fisher92}.
The transformation maps a correlation coefficient to a normal distribution, calculates the CI on the normal curve, and applies the reverse transformation to obtain the upper and lower bounds:
\begin{align*}
    z_r &= \textrm{arctanh}(r) \\
    r_u, r_\ell &= \textrm{tanh}\left(z_r \pm  z_{\alpha/2} \cdot c\; / \sqrt{n-b}\right)
\end{align*}
where $r$ is the correlation coefficient, $n$ is the number of observations, $z_{\alpha/2}$ is the critical value of a normal distribution, and $b$ and $c$ are constants.\footnote{
$b=3, 3, 4$ and $c=1, \sqrt{1+r^2/2}, \sqrt{.437}$ for Pearson, Spearman, and Kendall, respectively \citep{BonettWr00}.
}

Applying the Fisher transformation to calculate CIs for $\rhosys$ and $\rhosum$ is potentially problematic.
First, it assumes that the input variables are normally distributed \citep{BonettWr00}.
The metrics' scores and human annotations on the datasets that we experiment with are, in general, not normally distributed (see Appendix~\ref{appendix:normality}).
Thus, this assumption is violated, and we expect this is the case for other summarization datasets as well.
Second, it is not clear whether the transformation should be applied to the summary-level correlation since its final value is an average of correlations, which is not strictly a correlation.\footnote{
Correlation coefficients cannot be averaged because they are not additive in the arithmetic sense, however it is standard practice in summarization.
}

\subsection{Bootstrapping}
\label{sec:ci_bootstrapping}
A popular nonparametric method of calculating a CI is bootstrapping \citep{EfronTi93}.
Bootstrapping is a procedure that estimates the distribution of a test statistic by repeatedly sampling with replacement from the original dataset and calculating the test statistic on each sample.
Unlike the Fisher transformation, bootstrapping is a very flexible procedure that does not assume the data is normally distributed nor that the test statistic is a correlation, making it appropriate for summarization.

However, it is not clear how to perform bootstrap sampling for correlation levels.
Consider a more standard bootstrapped CI calculation for the mean accuracy of a question-answering model on a dataset with $k$ instances.
Since the mean accuracy is a function of the $k$ individual correct/incorrect labels, each bootstrap sample can be constructed by sampling with replacement from the original $k$ instances $k$ times.
In contrast, the correlation levels are functions of the matrices $X$ and $Z$, so each bootstrap sample should also be a pair of matrices of the same size that are sampled from the original data.

\begin{figure}
    \centering
    \includegraphics[width=\columnwidth]{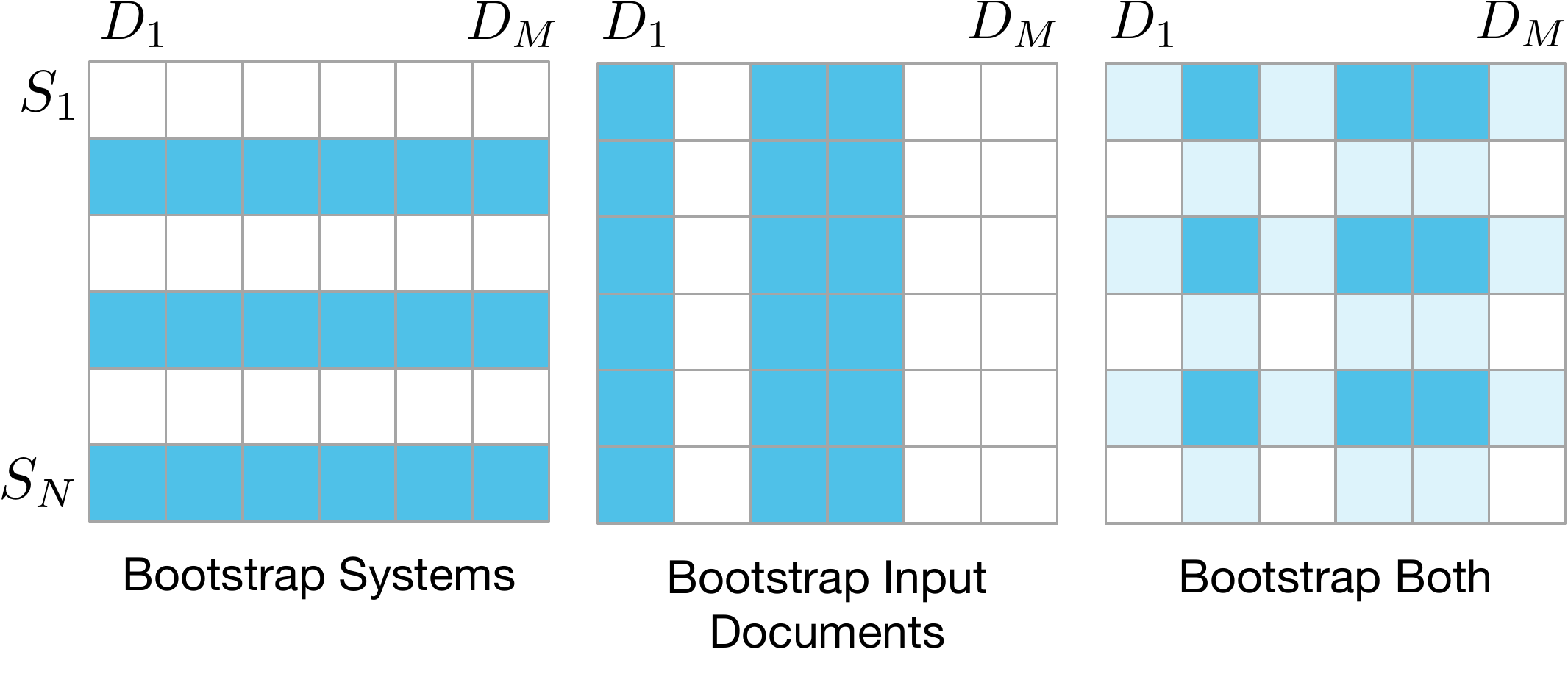}
    \caption{An illustration of the three methods for sampling matrices during bootstrapping.
    The dark blue color marks values selected by the sample.
    Only 3 system and input samples are shown here, when $N$ and $M$ are actually sampled with replacement.
    }
    \label{fig:sampling}
\end{figure}

There are at least three potential methods for sampling the matrices:
\begin{enumerate}
    \item \textsc{Boot-Systems:} Randomly sample with replacement $N$ systems from $\mathcal{S}$, then select the sampled system scores for all of the inputs.

    \item \textsc{Boot-Inputs:} Randomly sample with replacement $M$ inputs from $\mathcal{D}$, then select all of the system scores for the sampled inputs.
    
    \item \textsc{Boot-Both:} Randomly sample with replacement $M$ inputs from $\mathcal{D}$ and $N$ systems from $\mathcal{S}$, then select the sampled system scores for the sampled inputs.
\end{enumerate}
Once the samples are taken, the corresponding values from $X$ and $Z$ are selected to create the sampled matrices.
An illustration of each method is shown in Figure~\ref{fig:sampling}.

Each sampling method makes its own assumptions about the degrees of freedom in the sampling process that results in different interpretations of the corresponding CIs. \textsc{Boot-Inputs} assumes that there is only uncertainty on the inputs while the systems are held constant.
CIs derived from this sampling technique would express a range of values for the true correlation $\rho$ between $\mathcal{X}$ and $\mathcal{Z}$ for the \emph{specific} set of systems $\mathcal{S}$ and inputs from the same distribution as those in $\mathcal{D}$.
The opposite assumption is made for \textsc{Boot-Systems} (uncertainty in systems, inputs are fixed).
\textsc{Boot-Both}, which can be viewed as sampling systems followed by sampling inputs, assumes uncertainty on both the systems and the inputs.
Therefore the corresponding CI estimates $\rho$ for systems and inputs distributed the same as those in $\mathcal{S}$ and $\mathcal{D}$.

Algorithm~\ref{alg:ci} contains the pseudocode for calculating a CI via bootstrapping using the \textsc{Boot-Both} sampling method.
In \S\ref{sec:ci_simulations} we experimentally evaluate the Fisher transformation and the three bootstrap sampling methods, then analyze the CIs of several different metrics in \S\ref{sec:ci_experiments}.

\begin{algorithm}[t]
{
\small
\caption{Bootstrap Confidence Interval}
\label{alg:ci}
\hspace*{\algorithmicindent} \textbf{Input:} $X, Z \in \mathbb{R}^{N\times M}$, $k \in \mathbb{N}, \alpha \in [0, 1]$ \\
\hspace*{\algorithmicindent} \textbf{Output:} $(1-\alpha)\times 100\%$-confidence interval
\begin{algorithmic}[1]
\State samples $\gets$ an empty list
\For{$k$ iterations}
    \State $S$ $\gets$ samp. $\{1,\dots, N\}$ w/ repl. $N$ times
    \State $D$ $\gets$ samp. $\{1,\dots, M\}$ w/ repl. $M$ times
    \State $X_s, Z_s \gets$ empty $N \times M$ matrices
    \For{$(i, j) \in \{1, \dots, N\} \times \{1, \dots M\}$}
        \State $X_s[i, j] \gets X[S[i], D[j]]$
        \State $Z_s[i, j] \gets Z[S[i], D[j]]$
    \EndFor
    \State Append $r(X_s, Z_s)$ to samples
\EndFor
\State $\ell, u \gets (\alpha/2)\times 100$ and $(1-\alpha/2)\times 100$ percentiles of samples
\State \Return $\ell, u$
\end{algorithmic}
}
\end{algorithm}

\section{Significance Testing}
\label{sec:hypo}
Although CIs express the strength of the correlation between two metrics, they do not directly express whether one metric $\mathcal{X}$ correlates to another $\mathcal{Z}$ better than $\mathcal{Y}$ does due to their shared dependence on $\mathcal{Z}$.
This statistical analysis is performed by hypothesis testing.
The specific one-tailed hypothesis test we are interested in is:
\begin{align*}
    H_0 &: \rho(\mathcal{X}, \mathcal{Z}) - \rho(\mathcal{Y}, \mathcal{Z}) \leq 0 \\
    H_1 &: \rho(\mathcal{X}, \mathcal{Z}) - \rho(\mathcal{Y}, \mathcal{Z}) > 0
\end{align*}

\subsection{Williams' Test}
\label{sec:williams}
One method for hypothesis testing the difference between two correlations with a dependent variable that is used frequently to compare machine translation metrics is Williams' test \citep{Williams59}.
It uses the pairwise correlations between $X$, $Y$, and $Z$ to calculate a $t$-statistic and a corresponding $p$-value.\footnote{
The full equation is omitted for space.
See \citet{GrahamBa14} for details.
} Williams' test is frequently used to compare machine translation metrics' performances at the system-level \citep[among others]{MWFMB20}.

However, the test faces the same issues as the Fisher transformation: It assumes the input variables are normally distributed \citep{DunnCl71}, and it is not clear whether the test should be applied at the summary-level.

\subsection{Permutation Tests}
\label{sec:hypo_permutation}
Bootstrapping can be used to calculate a $p$-value in the form of a paired bootstrap test in which the sampling methods described in \S\ref{sec:ci_bootstrapping} can be used to resample new matrices from $X$, $Y$, and $Z$ in parallel (details omitted for space).
However, an alternative and closely related nonparametric hypothesis test is the permutation test \citep{Noreen89}.
Permutation tests tend to be used more frequently than paired bootstrap tests for hypothesis testing because they directly test whether any observed difference between two values is due to random chance.
In contrast, paired bootstrap tests indirectly reason about this difference by estimating the variance of the test statistic.

\begin{figure*}
    \centering
    \includegraphics[width=\textwidth]{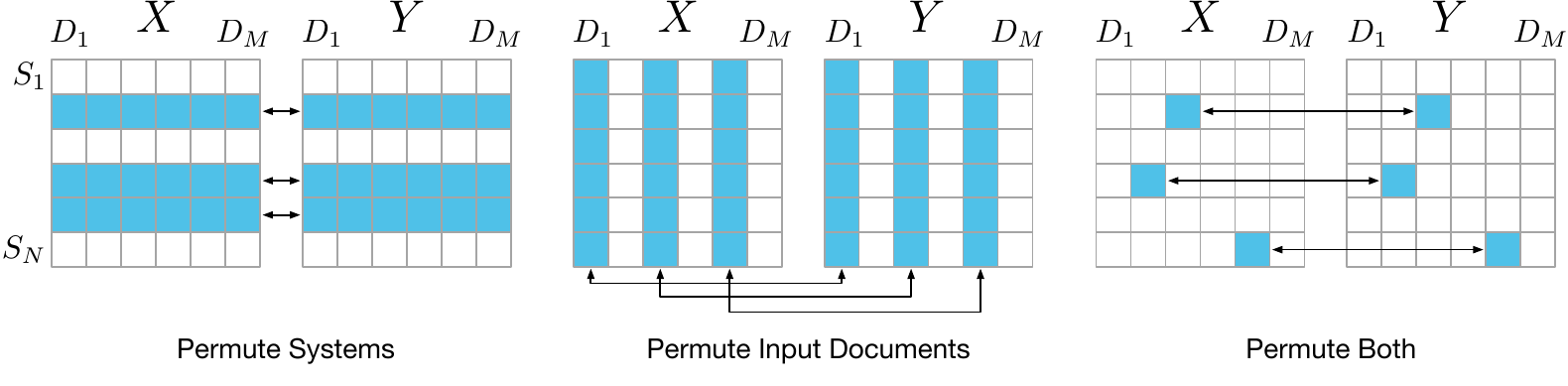}
    \caption{An illustration of the three permutation methods which swap system scores, document scores, or scores for individual summaries between $X$ and $Y$.
    }
    \label{fig:permutations}
\end{figure*}

Similarly to bootstrapping, a permutation test applied to two paired samples estimates the distribution of the test statistic under $H_0$ by calculating its value on new resampled datasets.
In contrast to bootstrapping, the resampled datasets are constructed by randomly permuting which sample each observation in a pair belongs to (i.e., resampling without replacement).
This relies on assuming the pair is exchangeable under $H_0$, which means $H_0$ is true for either sample assignment for the pair.
Then, the $p$-value is calculated as the proportion of times the test statistic across all possible permutations is greater than the observed value.
A significant $p$-value implies the observed test statistic is very unlikely to occur if $H_0$ were true, resulting in its rejection.
In practice, calculating the distribution of $H_0$ across all possible permutations is intractable, so it is instead estimated on a large number of randomly sampled permutations.\footnote{
    This is known as an approximate randomization test.
}

\begin{algorithm}[t]
{\small
\caption{Permutation Hypothesis Test}
\label{alg:permutation}
\hspace*{\algorithmicindent} \textbf{Input:} $X, Y, Z \in \mathbb{R}^{N\times M}$, $k \in \mathbb{N}, \alpha \in [0, 1]$
\hspace*{\algorithmicindent} \textbf{Output:} $p$-value
\begin{algorithmic}[1]
\State Standardize $X$ and $Y$
\State c $\gets$ 0
\State $\delta \gets r(X, Z) - r(Y, Z)$
\For{$k$ iterations}
    \State $X_s, Y_s \gets$ empty $N \times M$ matrices
    \For{$(i, j) \in \{1, \dots, N\} \times \{1, \dots, M\}$}
        \If{random Boolean is true} \Comment{swap}
            \State $X_s[i, j] \gets Y[i, j]$
            \State $Y_s[i, j] \gets X[i, j]$
        \Else \Comment{do not swap}
            \State $X_s[i, j] \gets X[i, j]$
            \State $Y_s[i, j] \gets Y[i, j]$
        \EndIf
    \EndFor
    \State $\delta_s \gets r(X_s, Z) - r(Y_s, Z)$
    \If{$\delta_s > \delta$}
        \State $c \gets c + 1$
    \EndIf
\EndFor
\State \Return $c / k$
\end{algorithmic}
}
\end{algorithm}

For example, a permutation test applied to testing the difference between two QA models' mean accuracies on the same dataset would sample a permutation by swapping the models' outputs for the same input.
Under $H_0$, the models' mean accuracies are equal, so randomly exchanging the outputs is not expected to change their means.
In the case of evaluation metrics, each permutation sample can be taken by randomly swapping the scores in $X$ and $Y$.
There are at least three ways of doing so:
\begin{enumerate}
    \item \textsc{Perm-Systems}: For each system, swap its scores for all inputs with probability 0.5.
    \item \textsc{Perm-Inputs}: For each input, swap its scores for all systems with probability 0.5.
    \item \textsc{Perm-Both}: For each summary, swap its scores with probability 0.5.
\end{enumerate}
To account for differences in scale, we standardize $X$ and $Y$ before performing the permutation.
Fig.~\ref{fig:permutations} contains an illustration of each method, and the pseudocode for a permutation test using the \textsc{Perm-Both} method is provided in Alg.~\ref{alg:permutation}.

Similarly to the bootstrap sampling methods, each of the permutation methods makes assumptions about the system and input document underlying distribution.
This results in different interpretations of how the tests' conclusions will generalize.
Since \textsc{Perm-Systems} randomly assigns system scores for all documents in $\mathcal{D}$ to either sample, we only expect the test's conclusion to generalize to a system distributed similarly to those in $\mathcal{S}$ evaluated on the \emph{specific} set of documents $\mathcal{D}$.
The opposite is true for \textsc{Perm-Inputs}.
The results for \textsc{Perm-Both} (which can be viewed as first swapping systems followed by swapping inputs) are expected to generalize for both systems and documents distributed similarly to those in $\mathcal{S}$ and $\mathcal{D}$.

In \S\ref{sec:power} we run a simulation to compare the different hypothesis testing approaches, then analyze the results of hypothesis tests applied to summarization metrics in \S\ref{sec:hypo_experiments}.

\section{Simulation Experiments}
\label{sec:simulations}
We run two sets of simulation experiments in order to determine which CI (\S\ref{sec:ci_simulations}) and hypothesis test (\S\ref{sec:power}) methods are most appropriate for summarization metrics.

The datasets used in the simulations are the multi-document summarization dataset TAC'08 \citep{DangOw08} and two subsets of the single-document summarization CNN/DM dataset \citep{NZSGX16} annotated by \citet{FKMSR21} and \citet{BGALN20}.
These datasets have $N=58/16/25$ summarization models and $M=48/100/100$ inputs, respectively.
The summaries were assigned overall responsiveness, relevance, or Lightweight Pyramid \citep{SGGRPBAD19} scores, respectively, by human annotators. The scores of the automatic metrics are correlated to these human annotations.

\subsection{Confidence Interval Simulation}
\label{sec:ci_simulations}
In practice, evaluation metrics are almost always used to score summaries produced by systems $\mathcal{S}'$ on inputs $\mathcal{D}'$ which are disjoint (or nearly disjoint) from and assumed to be distributed similarly to the data that was used to calculate the CI, $\mathcal{S}$ and $\mathcal{D}$.
It is still desirable to use the CI as an estimate of the correlation of a metric on $\mathcal{S}'$ and $\mathcal{D}'$, however this scenario violates assumptions made by some of the bootstraping sampling methods (e.g., \textsc{Boot-Systems} assumes that $\mathcal{D}$ is fixed).
This simulation aims to demonstrate the effect of violating these assumptions on the accuracy of the CIs.

\paragraph{Setup.}
The simulation works as follows.
The systems $\mathcal{S}$ and inputs $\mathcal{D}$ are each randomly partitioned into two equally sized disjoint sets $\mathcal{S}_A$, $\mathcal{S}_B$, $\mathcal{D}_A$, and $\mathcal{D}_B$.
Then the submatrices $X_A$, $Z_A$, $X_B$, and $Z_B$ are selected from $X$ and $Z$ based on the system and input partitions.
Matrices $X_A$ and $Z_A$ are used to calculate a 95\% CI using one of the methods described in \S\ref{sec:ci}, and then it is checked whether sample correlation $r(X_B, Z_B)$ is contained by the CI.
The entire procedure is repeated 1000 times, and the proportion of times the CI contains the sample correlation is calculated.

It is expected that a CI which generalizes well to the held-out data should contain the sample correlation 95\% of the time under the assumption that the data in $A$ and $B$ is distributed similarly.
The larger the difference from 95\%, the worse the CI is at estimating the correlation on the held-out data.

\begin{table}[t]
    \centering
    \begin{adjustbox}{width=\columnwidth}
    \begin{tabular}{ccccccccc}
        \toprule
        \multirow{2}{*}[-0.2em]{\makecell{\bf CI \\ \bf Method}} & \multicolumn{2}{c}{\bf TAC'08} & & \multicolumn{2}{c}{\bf Fabbri et al.} & & \multicolumn{2}{c}{\bf Bhandari et al.} \\
        \cmidrule{2-3} \cmidrule{5-6} \cmidrule{8-9}
         & $\rhosys$ & $\rhosum$ & & $\rhosys$ & $\rhosum$ & & $\rhosys$ & $\rhosum$ \\ 
        \midrule
Fisher & 0.72 & 1.00 & & 0.87 & 1.00 & & 0.85 & 1.00 \\
\textsc{Boot-Systems} & 0.76 & 0.72 & & 0.81 & 0.73 & & 0.80 & 0.72 \\
\textsc{Boot-Inputs} & 0.58 & 0.70 & & 0.70 & 0.73 & & 0.68 & 0.62 \\
\textsc{Boot-Both} & \bf 0.82 & \bf 0.92 & & \bf 0.98 & \bf 0.93 & & \bf 0.94 & \bf 0.88 \\
        \bottomrule
    \end{tabular}
    \end{adjustbox}
    \caption{The proportion of times the 95\% confidence interval for the true correlations $\rho$ of QAEval-F$_1$ calculated using Pearson contains the sample correlation of a held-out set of systems and inputs for the different methods of calculating confidence intervals.
    Values in bold are closest to 0.95 (and less than 1.0) and significantly different under a one-tailed difference of proportions $z$-test at $\alpha = 0.05$.
    }
    \label{tab:ci_simulation}
\end{table}

The results of the simulation calculated on TAC'08 and CNN/DM using both the Fisher transformation and the different bootstrap sampling methods to CIs for QAEval-F$_1$ \citep{DeutschBeRo20} are shown in Table~\ref{tab:ci_simulation}.\footnote{
    The Fisher transformation was directly applied to the averaged summary-level correlation.
}

\paragraph{\textsc{Boot-Both} generalizes the best.}
Among the bootstrap methods, \textsc{Boot-Both} produces CIs that come closest to the ideal 95\% rate.
Any deviations from this number reflect that the assumption that all of the inputs and systems are distributed similarly is not true, but overall violating this assumption does not have a major impact.

The other bootstrap methods, which sample only systems or inputs, captures the correlation on the held-out data far less than 95\% of the time.
For instance, the CIs for $\rhosys$ on \citet{BGALN20} only successfully estimate the held-out correlation on 80\% and 68\% of trials.
This means that a 95\% CI calculated using \textsc{Boot-Inputs} is actually only a 68\% CI on the held-out data.
This pattern is the same across the different correlation levels and datasets.
The lower values for only sampling inputs indicates that more variance comes from the systems rather than the inputs.

\paragraph{Fisher analysis.}
The Fisher transformation at the system-level creates CIs that generalize worse than \textsc{Boot-Both}.
The summary-level CI captures the held-out sample correlation 100\% of the time, implying that the CI width is too large to be useful.
We believe this is due to the fact that as the absolute value of $r(X, Z)$ decreases, the width of the Fisher CI increases.
Summary-level correlations are lower than system-level correlations (see \S\ref{sec:ci_experiments}), and therefore Fisher results in a worse CI estimate at the summary-level.

\paragraph{Conclusion.}
This experiment presents strong evidence that violating the assumptions that either the systems/inputs are fixed or that the data is normally distributed does result in worse CIs.
Hence, the \textsc{Boot-Both} method provides the most accurate CIs for scenarios in which summarization metrics are frequently used.

\subsection{Power Analysis}
\label{sec:power}
The power of a hypothesis test is the probability of accepting the alternative hypothesis given that it is actually true (equal to $1.0$ -- the type-II error rate).
It is desirable to have as high of a power as possible in order to avoid missing a significant difference between metrics.
This simulation estimates the power of each of the hypothesis tests.

\paragraph{Setup.}
Measuring power requires a scenario in which it is known that $\rho$ is greater for one metric than another (i.e., $H_1$ is true).
Since this is not known to be true for any pair of proposed evaluation metrics, we artificially create such a scenario by adding randomness to the calculation of ROUGE-1.\footnote{
    We use the recall variant of ROUGE for experiments on TAC'08 and \citet{BGALN20} and the F$_1$ variant on \citet{FKMSR21} throughout the paper.
}
We define $\mathcal{R}_k$ to be ROUGE-1 calculated using a random $k\%$ of the candidate summary's tokens.
We assume that since $\mathcal{R}_k$ only evaluates a summary with $k\%$ of its tokens, it is quite likely that it is a worse metric than standard ROUGE-1 for $k < 100$.

To estimate the power, we score summaries with ROUGE-1 and $\mathcal{R}_k$ for different $k$ values and count how frequently each hypothesis test rejects $H_0$ in favor of identifying ROUGE-1 as a superior metric.
This trial is repeated 1000 times, and the proportion of significant results is the estimate of the power.

Since the various hypothesis tests make different assumptions about whether the systems and inputs are fixed or variable, it is not necessarily fair to directly compare their powers.
Because the assumptions of \textsc{Boot-Both} and \textsc{Perm-Both} most closely align with the typical use case of summarization, we compare their powers.
We additionally include Williams' test because it is frequently used for machine translation metrics and it produces interesting results, discussed below.

\begin{figure}
    \centering
    \includegraphics[width=\columnwidth]{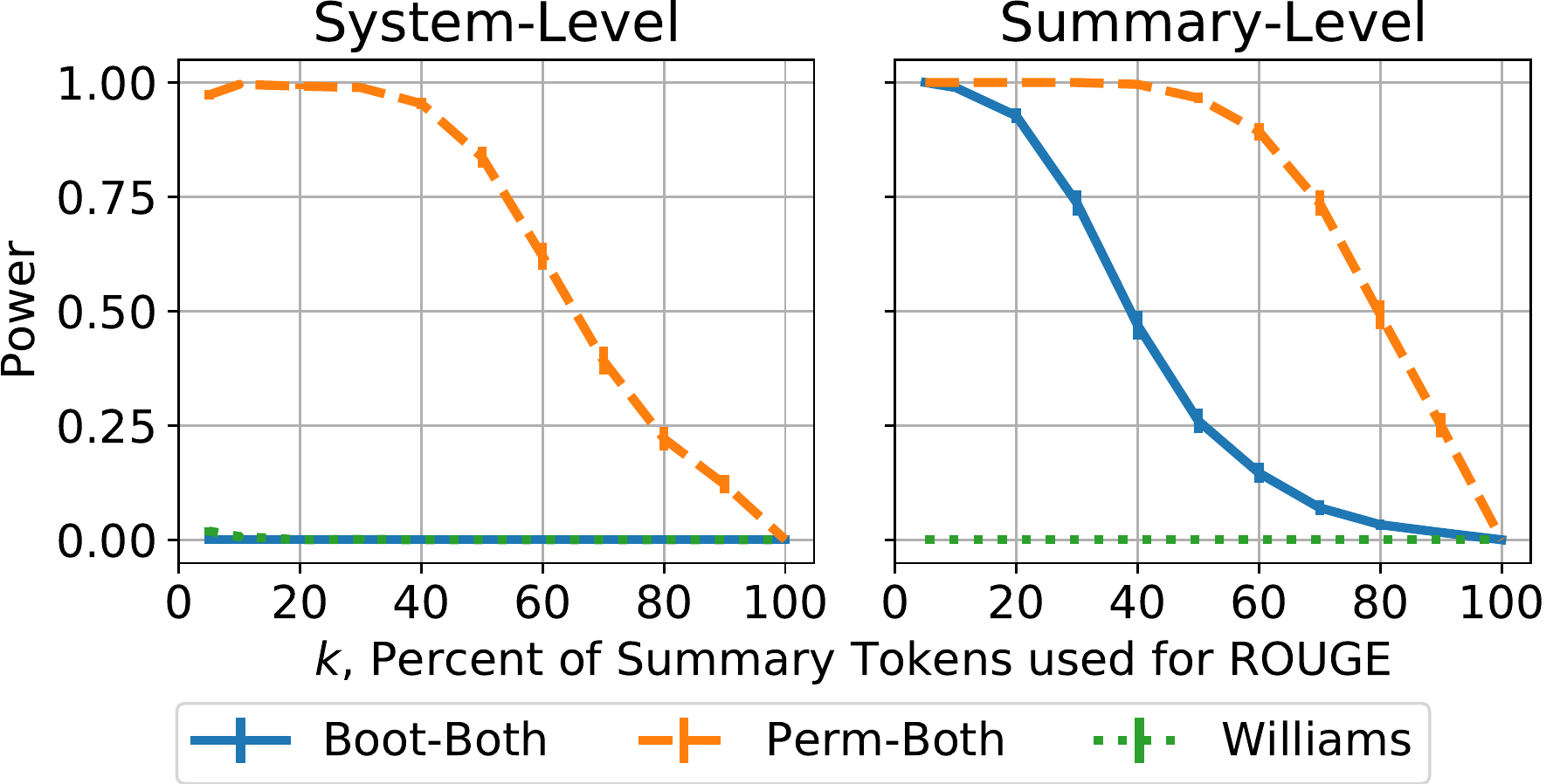}
    \caption{The system- and summary-level Pearson estimates of the power of the \textsc{Boot-Both}, \textsc{Perm-Both}, and Williams hypothesis test methods calculated on the annotations from \citet{FKMSR21}.
    The power for \textsc{Boot-Both} and Williams at the system-level is $\approx 0$ for all values.}
    \label{fig:power}
\end{figure}

\paragraph{\textsc{Perm-Both} has the highest power.}
Fig.~\ref{fig:power} plots the power curves for various values of $k$ on the CNN/DM annotations by \citet{FKMSR21}.
We find that \textsc{Perm-Both} has the highest power among the three tests for all values of $k$.
As $k$ approaches $100\%$, the difference between ROUGE-1 and $\mathcal{R}_k$ becomes smaller and harder to detect, thus the power for all methods approaches 0.

\textsc{Boot-Both} has lower power than \textsc{Perm-Both} both at the summary-level and system-level, in which it is near 0.
This result is consistent with permutation tests being more useful for hypothesis testing than their bootstrapping counterparts.
We believe the power differences in both levels are due to the variance of the two correlation levels.
As we observe in \S\ref{sec:ci_experiments}, the system-level CIs have significantly larger variance than at the summary-level, making it harder for the paired bootstrap to reject the system-level $H_0$.

\begin{figure*}
    \centering
    \includegraphics[width=1.0\textwidth]{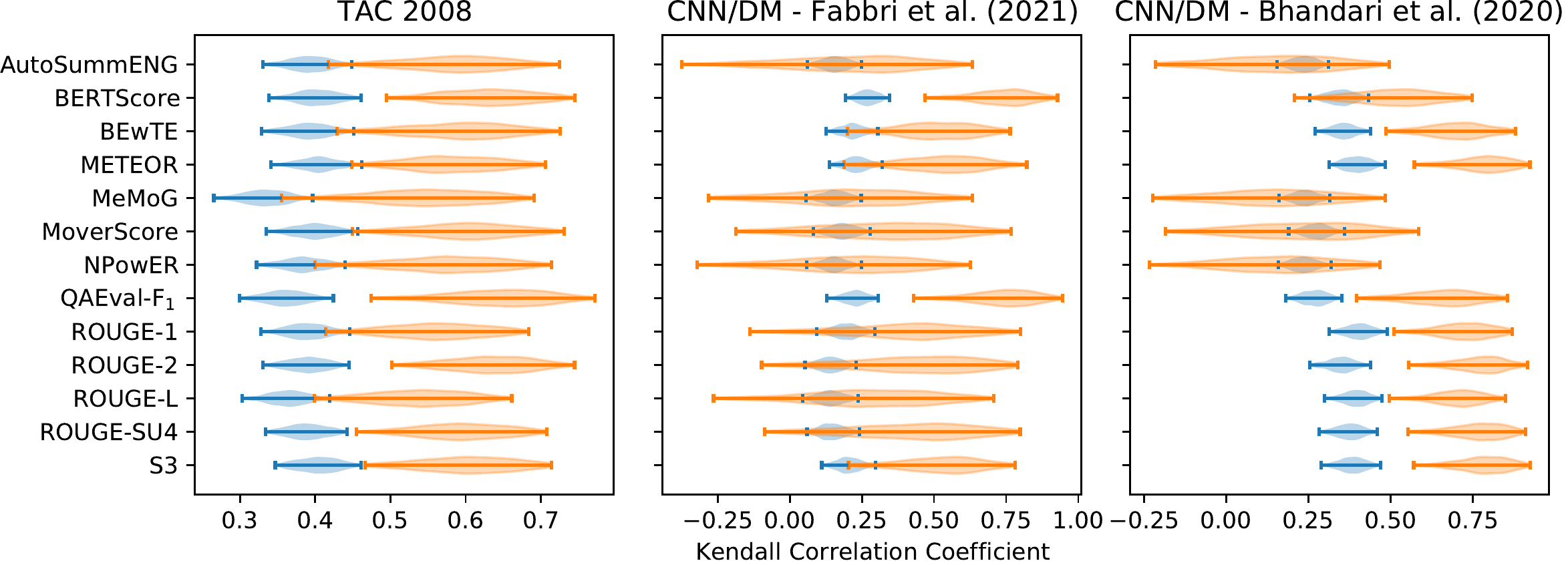}
    \caption{The 95\% confidence intervals for $\rhosum$ (blue) and $\rhosys$ (orange) calculated using Kendall's correlation coefficient on TAC'08 (left) and CNN/DM summaries (middle, \citet{FKMSR21}; right, \citet{BGALN20}) are rather large, reflecting the uncertainty about how well these metrics agree with human judgments of summary quality.}
    \label{fig:ci}
\end{figure*}

\paragraph{Williams' test has low power.}
Interestingly, the power of Williams' test for all $k$ is $\approx 0$, implying the test never rejects $H_0$ in this simulation.
This is surprising because Williams' test is frequently used to compare machine translation metrics at the system-level and does find differences between metrics.
We believe this is due to the strength of the correlations of ROUGE-1 to the ground-truth judgments as follows.

The $p$-value calculated by Williams is a function of the pairwise correlations of $X$, $Y$, and $Z$ and the number of observations.
The closer both $r(X, Z)$ and $r(Y, Z)$ are to 0, the higher the $p$-value.
The correlation of ROUGE-1 in this simulation is around 0.6 and 0.3 at the system- and summary-levels.
In contrast, the system-level correlations for the metrics submitted to the Workshop on Machine Translation (WMT) 2019's metrics shared task for de-en are on average 0.9 \citep{MWBG19}.
Among the 231 possible pairwise metric comparisons in WMT'19 for de-en, Williams' test yields 81 significant results.
If the correlations are shifted to have an average value of 0.6, only 3 significant results are found.
Thus we conclude that Williams' test's power is worse for detecting differences between lower correlation values.

Because this simulation is performed with summarization metrics on a real summarization dataset, we believe it is faithful enough to a realistic scenario to conclude that Williams' test does indeed have low power when applied to summarization metrics.
However, we do not expect Williams' test to have 0 power when used to detect differences between machine translation metrics.

\paragraph{Conclusion.}
Since \textsc{Perm-Both} has the best statistical power at both the system- and summary-levels, we recommend it for hypothesis testing the difference between summarization metrics.
\section{Summarization Analysis}
We run two experiments that calculate CIs (\S\ref{sec:ci_experiments}) and run hypothesis tests (\S\ref{sec:hypo_experiments}) for many different summarization metrics on the TAC'08 and CNN/DM datasets (\S\ref{sec:simulations}).
Each experiment also includes an analysis which discusses the implications of the results for the summarization community.

The metrics used for experimentation are the following:
AutoSummENG \citep{GKVS08},
BERTScore \citep{ZKWWA20},
BEwT-E \citep{TratzHo08},
METEOR \citep{DenkowskiLa14},
MeMoG \citep{GiannakopoulosKa10},
MoverScore \citep{ZPLGME19},
NPowER \citep{GiannakopoulosKa13},
QAEval \citep{DeutschBeRo20},
ROUGE \citep{Lin04},
and S$^3$ \citep{PeyrardBoGu17}.
We use the metrics' implementations in the SacreROUGE library \citep{DeutschRo20}.

\subsection{Confidence Intervals}
\label{sec:ci_experiments}
Fig.~\ref{fig:ci} shows the 95\% CIs calculated via \textsc{Boot-Both} for $\rhosum$ and $\rhosys$ for each metric calculated using Kendall's $\tau$.
Since ROUGE is the most commonly used metric, the following discussion will mostly focus on its results, however the conclusions largely apply to other metrics as well.

\paragraph{Confidence intervals are large.}
The most apparent observation is that the CIs are rather large, especially for $\rhosys$.
The ROUGE-2 $\rhosys$ CIs are $[.49, .74]$ for TAC'08 and $[-.09, .84]$ on CNN/DM using the annotations from \citet{FKMSR21}.
The wide range of values demonstrates that there is a large amount of uncertainty around how precise the correlations reported in the literature truly are.

The size of the CIs has serious implications for how trustable existing automatic evaluations are.
Since Kendall's $\tau$ is a function of the number of pairs of systems in which the automatic metric and ground-truth agree on their rankings, the metrics' CIs can be translated to upper- and lower-bounds on the number of incorrect rankings.
Specifically, ROUGE-2's system-level CI on \citet{FKMSR21} implies it incorrectly ranks systems with respect to humans 9-54\% of the time.
This means that potentially more than half of the time ROUGE ranks one summarization model higher than another on CNN/DM, it is wrong according to humans, a rather surprising result.
However, it is consistent with similar findings by \citet{RCDN13}, who estimated the same result to be around 37\% for top-performing systems on TAC 2008-2011.

We suspect that the true ranking accuracy of ROUGE (as well as the other metrics) is not likely to be at the extremes of the confidence interval due to the distribution of the bootstrapping samples shown in Fig.~\ref{fig:ci}.
However, this experiment highlights the uncertainty around how well automatic metrics replicate human annotations of summary quality.
An improved ROUGE score does not necessarily mean a model produces better summaries.
Likewise, not improving ROUGE should not disqualify a model from further consideration.
Consequently, researchers should rely less heavily on automatic metrics for determining the quality of summarization models than they currently do.
Instead, the community needs to develop more robust evaluation methodologies, whether it be task-specific downstream evaluations or faster and cheaper human evaluation.

\paragraph{Comparing CNN/DM annotations.}
The CIs calculated on the annotations by \citet{BGALN20} are in general higher and more narrow than on \citet{FKMSR21}.
We believe this is due to the method of selecting the summaries to be annotated for each of the datasets.
\citet{BGALN20} selected summaries based on a stratified sample of automatic metric scores, whereas \citet{FKMSR21} selected summaries uniformly at random.
Therefore, the summaries in \citet{BGALN20} are likely easier to score (due to a mix of high- and low-quality summaries) and are less representative of the real data distribution than those in \citet{FKMSR21}.

\begin{figure*}
    \centering
    \includegraphics[width=1.0\textwidth]{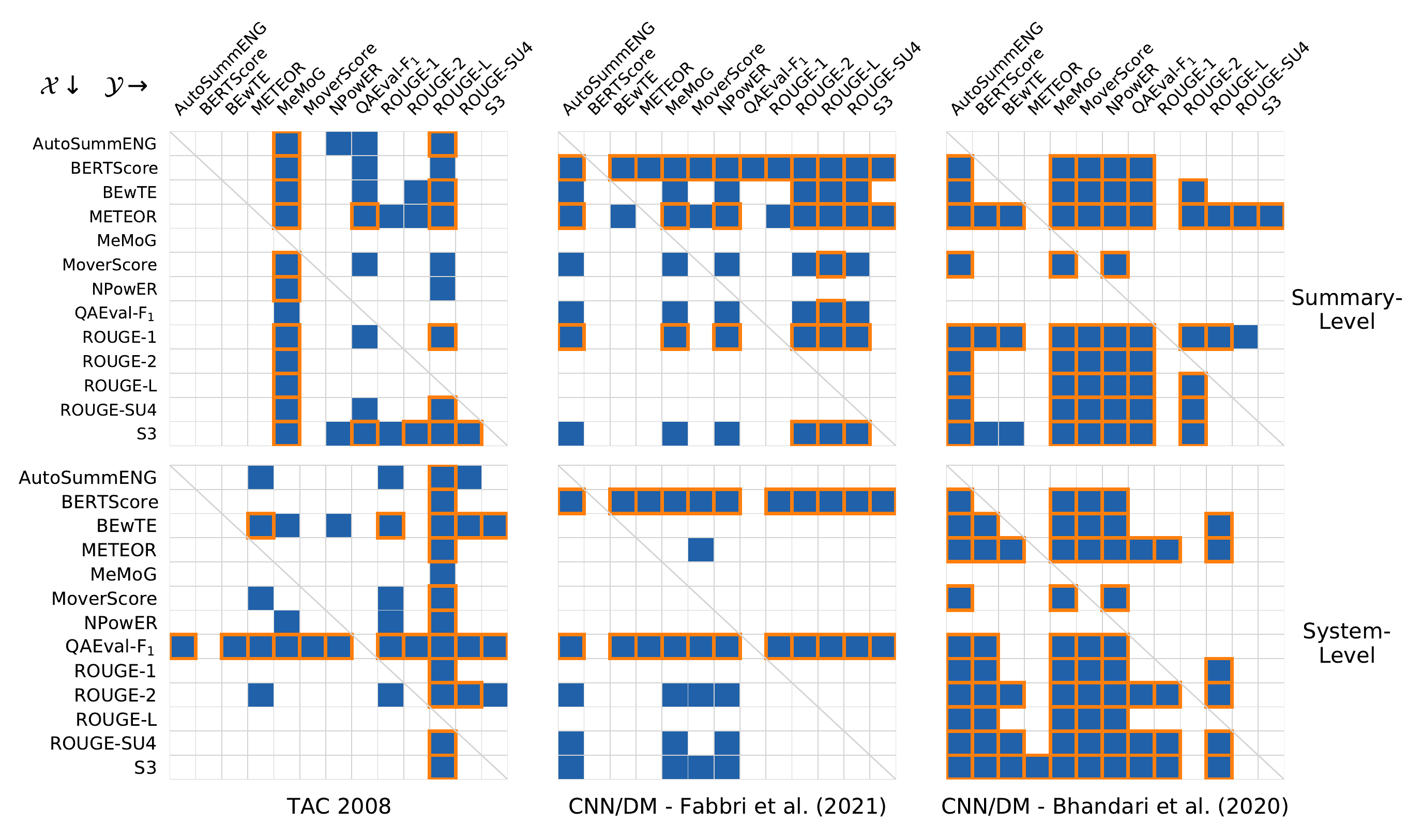}
    \caption{The results of running the \textsc{Perm-Both} hypothesis test to find a significant difference between metrics' Pearson correlations.
    A blue square means the test returned a significant $p$-value at $\alpha = 0.05$, indicating the row metric has a higher correlation than the column metric.
    An orange outline means the result remained significant after applying the Bonferroni correction.
    }
    \label{fig:hypo}
\end{figure*}

\subsection{Hypothesis Testing}
\label{sec:hypo_experiments}
Although nearly all of the CIs for the metrics are overlapping, this does not necessarily mean that no metric is statistically better than another since the differences between two metrics' correlations could be significant.

In Fig.~\ref{fig:hypo}, we report the $p$-values for testing $H_0: \rho(\mathcal{X}, \mathcal{Z}) - \rho(\mathcal{Y}, \mathcal{Z}) \leq 0$ using the \textsc{Perm-Both} permutation test at the system- and summary-levels on TAC'08 and CNN/DM for all possible metric combinations (see \citet{AKSR20} for a discussion about how to interpret $p$-values).
The Bonferroni correction \citep[which lowers the significance level for rejecting each individual null hypothesis such that the probability of making one or more type-I errors is bounded by $\alpha$;][]{Bonferroni36,DBBR17} was applied to test suites grouped by the $\mathcal{X}$ metric at $\alpha = 0.05$.\footnote{
    A version of the results when the correction is applied to $p$-values grouped by the dataset and correlation level pair is included in Appendix~\ref{sec:bonferroni_full}.
}
A significant result means that we conclude that $\rho(\mathcal{X}, \mathcal{Z}) > \rho(\mathcal{Y}, \mathcal{Z})$.

The metrics which are identified as being statistically superior to others at the system-level on TAC'08 and CNN/DM using the annotations from \citet{FKMSR21} are QAEval and BERTScore.
Although they are statistically indistinguishable from each other, QA\-Eval does improve over more metrics than BERTScore does on TAC'08.
At the summary-level, BERTScore has significantly better results than all other metrics.
Overall, none of the other metrics consistently outperform all variants of ROUGE.
Results using either the Spearman or Kendall correlation coefficients are largely consistent with Fig.~\ref{fig:hypo}, although QA\-Eval no longer improves over some metrics, such as ROUGE-2, at the system-level on TAC'08.

The results on the CNN/DM annotations provided by \citet{BGALN20} are less clear.
The ROUGE variants appear to perform well, a conclusion also reached by \citet{BGALN20}.
The hypothesis tests also find that S3 is statistically better than most other metrics.
S3 scores systems using a learned combination of features which includes ROUGE scores, likely explaining this result.
Similarly to the CI experiment, the results on the annotations provided by \citet{BGALN20} and \citet{FKMSR21} are rather different, potentially due to differences in how the datasets were sampled.
\citet{FKMSR21} uniformly sampled summaries to annotate, whereas \citet{BGALN20} sampled them based on their approximate quality scores, so we believe the dataset of \citet{FKMSR21} is more likely to reflect the real data distribution.

\section{Limitations}
The large widths of the CIs in \S\ref{sec:ci_experiments} and the lack of some statistically significant differences between metrics in \S\ref{sec:hypo_experiments} are directly tied to the size of the datasets that were used in our analyses.
However, to the best of our knowledge, the datasets we used are some of the largest available with annotations of summary quality.
Therefore, the results presented here are our best efforts at accurately measuring the metrics' performances with the data available.
If we had access to larger datasets with more summaries labeled across more systems, we suspect that the scores of the human annotators and automatic metrics would stabilize to the point where the CI widths would narrow and it would be easier to find significant differences between metrics.

Although it is desirable to have larger datasets, collecting them is difficult because obtaining human annotations of summary quality is expensive and prone to noise.
Some studies report having difficulty obtaining high-quality judgments from crowdworkers \citep{GillickLi10,FKMSR21}, whereas others have been successful using the crowdsourced Lightweight Pyramid Score \citep{SGGRPBAD19}, which was used in \citet{BGALN20}.

Then, it is unclear how well our experiments' conclusions will generalize to other datasets with different properties, such as documents coming from different domains or different length summaries.
The experiments in \citet{BGALN20} show that metric performance depends on which dataset you use to evaluate, whether it be TAC or CNN/DM, which is supported by our results.
However, our experiments also show variability in performance within the same dataset when using different quality annotations (see the differences in results between \citet{FKMSR21} and \cite{BGALN20}).
Clearly, more research needs to be done to understand how much of these changes in performance is due to differences in the properties of the input documents and summaries versus how the summaries were annotated.
\section{Related Work}
\paragraph{Summarization}
CIs and hypothesis testing were applied for summarization evaluation metrics over the years in a relatively inconsistent manner -- if at all.
To the best of our knowledge, the only instances of calculating CIs for summarization metrics is at the system-level using a bootstrapping procedure equivalent to \textsc{Boot-Systems} \citep{RankelCoSc12,DavisCoSc12}.
Some works do perform hypothesis testing, but it is not clear which statistical test was run \citep{TratzHo08, GKVS08}.
Others report whether or not the correlation itself is significantly different from 0 \citep{Lin04}, which does not quantify the strength of the correlation nor allow for comparisons.
Some studies apply Williams' test to compare summarization metrics.
For instance, \citet{Graham15} use it to compare BLEU \citep{PRWZ02} and several variants of ROUGE, and \citet{BGALN20} compares several different metrics at the system-level.
However, our experiments demonstrated in \S\ref{sec:power} that Williams' test has lower power than the suggested methods due to the lower correlation values.

As an alternative to comparing metrics' correlations, \citet{OCDN12} argue for comparison based on the number of system pairs in which both human judgments and metrics agree on statistically significant differences between the systems, a metric also used in the TAC shared-task for summarization metrics \citep[][\emph{i.a.}]{DangOw09}.
This can be viewed similarly to Kendall's $\tau$ in which only statistically significant differences between systems are counted as concordant.
However, the differences in discriminative power across metrics was not statistically tested itself.

More broadly in evaluating summarization systems, \citet{RCSO11} argue for comparing the performance of summarization models via paired $t$-tests or Wilcoxon signed-rank tests \citep{Wilcoxon92}.
They demonstrate these tests have more power than the equivalent unpaired test when used to separate human and model summarizers.

\paragraph{Machine Translation}
The summarization and machine translation (MT) communities face the same problem of developing and evaluating automatic metrics to evaluate the outputs of models.
Since 2008, the Workshop on Machine Translation (WMT) has run a shared-task for developing evaluation metrics \citep[among others]{MWFMB20}.
Although the methodology has changed over the years, they have converged on comparing metrics' system-level correlations using Williams' test \citep{GrahamBa14}.
Since Williams' test assumes the input data is normally distributed and our experiments show it has low power for summarization, we do not recommend it for comparing summarization metrics.
However, human annotations for MT are standardized to be normally distributed, and the metrics have higher correlations to human judgments, thus Williams' test will probably have higher power when applied to MT metrics.
Nevertheless, the methods proposed in this work can be directly applied to MT metrics as well.

\section{Conclusion}
In this work, we proposed several different methods for estimating CIs and hypothesis testing for summarization evaluation metrics using resampling methods.
Our simulation experiments demonstrate that assuming variability in both the systems and input documents leads to the best generalization for CIs and that permutation-based hypothesis testing has the highest statistical power.
Experiments on several different evaluation metrics across three datasets demonstrate high uncertainty in how well metrics correlate to human judgments and that QA\-Eval and BERTScore do achieve higher correlations than ROUGE in some settings.

\section*{Acknowledgments}
The authors would like to thank Lyle Ungar, Daniel Khashabi, Eyal Ben David, and the anonymous reviewers for their valuable feedback on our work.

This work was partly supported by a a Focused Award from Google, by contracts FA8750-19-2-1004 and FA8750-19-2-0201 with the US Defense Advanced Research Projects Agency (DARPA), and by the Office of the Director of National Intelligence (ODNI), Intelligence Advanced Research Projects Activity (IARPA), via IARPA Contract No. 2019-19051600006 under the BETTER Program.
The views and conclusions contained herein are those of the authors and should not be interpreted as necessarily representing the official policies, either expressed or implied, of ODNI, IARPA, DARPA, the Department of Defense, or the U.S. Government. The U.S. Government is authorized to reproduce and distribute reprints for governmental purposes notwithstanding any copyright annotation therein.

\bibliography{new_ccg,new_cited,ccg,cited}
\bibliographystyle{acl_natbib}

\appendix
\begin{table}
    \centering
    \begin{adjustbox}{width=\columnwidth}
    \begin{tabular}{ccccccccc}
        \toprule
        \multirow{2}{*}{\textbf{Metric}} & \multicolumn{2}{c}{\textbf{TAC'08}} & & \multicolumn{2}{c}{\textbf{Fabbri et al.}} & & \multicolumn{2}{c}{\textbf{Bhandari et al.}} \\
        \cmidrule{2-3} \cmidrule{5-6} \cmidrule{8-9}
        & $r_\textsc{Sum}$ & $r_\textsc{Sys}$ & & $r_\textsc{Sum}$ & $r_\textsc{Sys}$ & & $r_\textsc{Sum}$ & $r_\textsc{Sys}$ \\
        \midrule
Resp/Rel/Pyr & 100.0 & 0.00 &   & 32.0 & 0.52 &   & 75.0 & 0.84\\
AutoSummENG & 18.8 & 0.26 &   & 33.0 & 0.01 &   & 28.0 & 0.55\\
MeMoG & 37.5 & 0.53 &   & 33.0 & 0.01 &   & 28.0 & 0.55\\
NPowER & 29.2 & 0.36 &   & 33.0 & 0.01 &   & 28.0 & 0.55\\
BERTScore & 35.4 & 0.00 &   & 26.0 & 0.15 &   & 28.0 & 0.18\\
BEwTE & 22.9 & 0.06 &   & 37.0 & 0.00 &   & 33.0 & 0.68\\
METEOR & 27.1 & 0.15 &   & 27.0 & 0.00 &   & 30.0 & 0.61\\
MoverScore & 47.9 & 0.25 &   & 35.0 & 0.00 &   & 31.0 & 0.50\\
QAEval-F$_1$ & 58.3 & 0.00 &   & 40.0 & 0.01 &   & 45.0 & 0.21\\
ROUGE-1 & 33.3 & 0.06 &   & 32.0 & 0.00 &   & 30.0 & 0.91\\
ROUGE-2 & 31.2 & 0.71 &   & 34.0 & 0.00 &   & 61.0 & 0.62\\
ROUGE-L & 25.0 & 0.13 &   & 26.0 & 0.13 &   & 37.0 & 0.12\\
ROUGE-SU4 & 29.2 & 0.44 &   & 32.0 & 0.00 &   & 44.0 & 0.84\\
S3 & 20.8 & 0.32 &   & 26.0 & 0.00 &   & 47.0 & 0.66\\
  \bottomrule
    \end{tabular}
    \end{adjustbox} 
    \caption{For $\rsys$ the $p$-value of the Shapiro-Wilk test.
    For $\rsum$, the percent of the per-input document tests which had a significant result at $\alpha = 0.05$.
    A significant $p$-value means $H_0$ (the data is distributed normally) is rejected.
    For $\rsum$, the larger the percentage the more the data appears to be not normally distributed.}
    \label{tab:normality}
\end{table}

\section{Normality Testing}
\label{appendix:normality}

To understand if the normality assumption holds for summarization data we ran the Shapiro-Wilk test for normality \citep{ShapiroWi65}, which was reported to have the highest power out of several alternatives \citep{RazaliWa11,DBSR18,DPSR20}.
The results of the tests for the ground-truth responsiveness scores and automatic metrics are in Table~\ref{tab:normality}.
Most of the $p$-values are significant, i.e., applying a statistical test which assumes normality is incorrect in general.

\section{Extended Bonferroni Correction}
\label{sec:bonferroni_full}
\begin{figure*}[h!]
    \centering
    \includegraphics[width=1.0\textwidth]{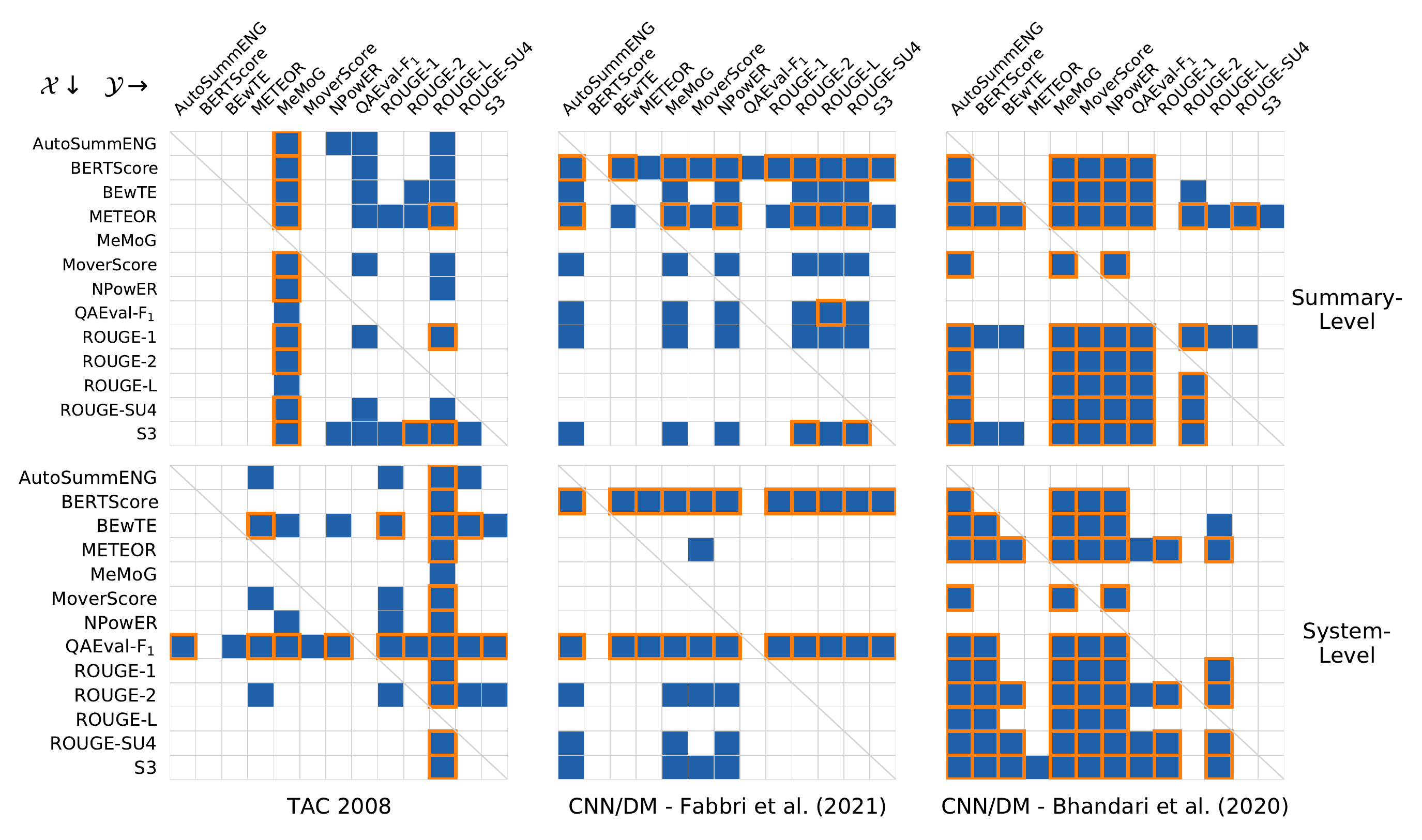}
    \caption{The results of running the \textsc{Perm-Both} hypothesis test to find a significant difference between metrics' Pearson correlations with the Bonferroni correction applied per dataset and correlation level pair instead of per metric (as in Fig.~\ref{fig:hypo}).
    A blue square means the test returned a significant $p$-value at $\alpha = 0.05$, indicating the row metric has a higher correlation than the column metric.
    An orange outline means the result remained significant after applying the Bonferroni correction.
    }
    \label{fig:hypo_full_table}
\end{figure*}

Fig.~\ref{fig:hypo_full_table} contains the results from the pairwise hypothesis tests (\S\ref{sec:hypo_experiments}) when then Bonferroni correction is applied to set of $p$-values grouped by the dataset and correlation level pair instead of each dataset, correlation level, and metric shown in Fig.~\ref{fig:hypo}.
The results are overall very similar with only a handful of results now becoming not significant.

\end{document}